\pdfoutput=1

\documentclass[11pt]{article}

\usepackage[]{acl}

\usepackage{times}
\usepackage{latexsym}
\usepackage{amsmath,amssymb}

\usepackage[T1]{fontenc}

\usepackage[utf8]{inputenc}

\usepackage{microtype}

\usepackage{listings}
\usepackage{graphicx}
\usepackage{float}
\usepackage{array}
\usepackage{booktabs}
\usepackage{verbatim}

\newlength\tbspace
\newcolumntype{L}{c<{\hspace{\tbspace}}}

\title{
Feature Interactions Reveal Linguistic Structure in Language Models
}

\author{Jaap Jumelet\hspace{1cm}Willem Zuidema \\
  Institute for Logic, Language and Computation \\
  University of Amsterdam \\
  \texttt{\{j.w.d.jumelet, w.zuidema\}@uva.nl} \\
}

\begin{document}
\maketitle
\begin{abstract}
We study \textit{feature interactions} in the context of \emph{feature attribution} methods for post-hoc interpretability. 
In interpretability research, getting to grips with feature interactions is increasingly recognised as an important challenge, because interacting features are key to the success of neural networks. Feature interactions allow a model to build up hierarchical representations for its input, and might provide an ideal starting point for the investigation into linguistic structure in language models.
However, uncovering the exact role that these interactions play is also difficult, and a diverse range of interaction attribution methods has been proposed.
In this paper, we focus on the question which of these methods most \textit{faithfully} reflects the inner workings of the target models.
We work out a \emph{grey box} methodology, in which we train models to perfection on a formal language classification task, using PCFGs. 
We show that under specific configurations, some methods are indeed able to uncover the grammatical rules acquired by a model. 
Based on these findings we extend our evaluation to a case study on language models, providing novel insights into the linguistic structure that these models have acquired.\footnote{All code and data is available here: \url{https://github.com/jumelet/fidam-eval}}
\end{abstract}


\section{Introduction}\label{sec:introduction}

\begin{figure*}[ht]
    \centering
    \includegraphics[width=\textwidth]{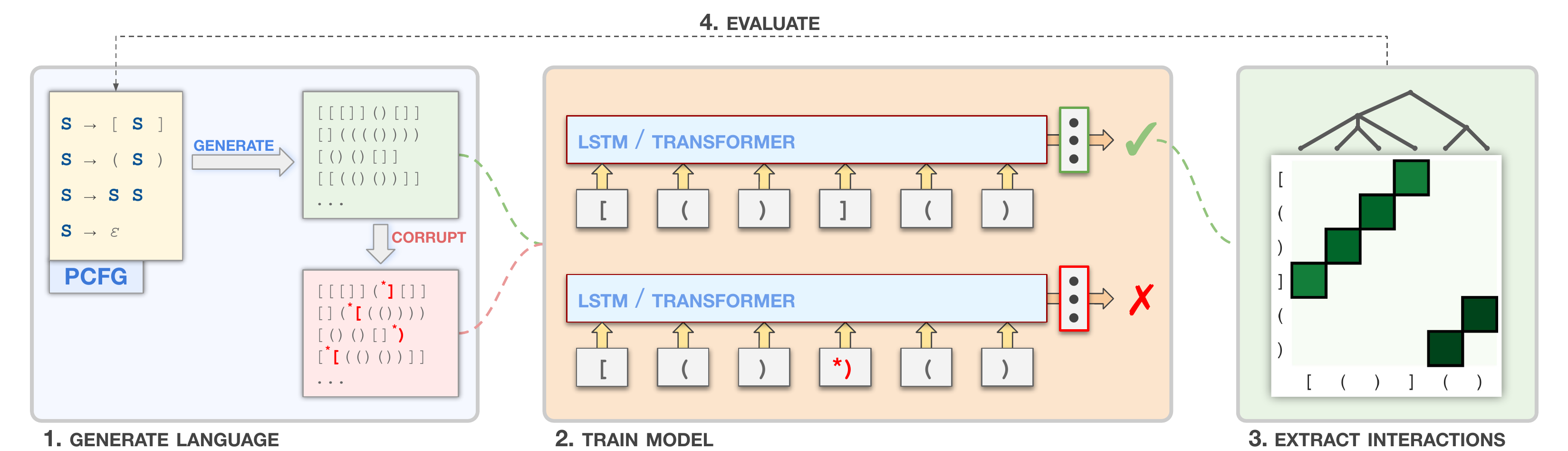}
    \caption{
    We generate a corpus based on a PCFG, and create negative examples by corrupting the generated corpus.
    Next, we train a neural model to predict whether a string is well-formed, forcing the model to obtain a comprehensive understanding of the rules of the language.
    Then, we extract the internal interactions by using FIDAMs described in \S\ref{sec:interactions}, allowing us to directly evaluate the grammatical knowledge of the neural model.
    }
    \label{fig:pipeline}
\end{figure*}

Feature attribution methods (FAMs) are a popular family of tools for explaining the behaviour of deep learning models, by explaining a prediction in terms of contributions of individual features \citep{DBLP:conf/kdd/Ribeiro0G16, DBLP:conf/nips/LundbergL17}. There are many such methods proposed, and mathematical results (such as axiomatic approaches based on game theory) and theoretical frameworks (such as \citet{DBLP:journals/jmlr/CovertLL21}'s `Explaining by Removing') are starting to offer a good understanding of how different methods relate to one another.

However, there are also some important shortcomings. 
%
Perhaps most importantly, popular FAMs mostly ignore the existence of interactions between the effects of features on the prediction. This is problematic, because \textbf{Feature Interactions} are widely seen as a major factor in the success of neural networks \citep{goodfellow2016}. 
%
%
This is all the more important in domains such as language and music processing, because feature interactions allow neural networks to model hierarchical representations of their input, which is considered a key design feature of language and music. 
To address these shortcomings, there is now 
an emerging literature on \textbf{feature interaction detection and attribution methods} (FIDAMs) that explain model predictions in terms of interacting features \citep{DBLP:conf/nips/TsangR020, DBLP:journals/jmlr/JanizekSL21}.



However, 
assessing the faithfulness of FIDAMs is even more challenging than assessing the faithfulness of feature attribution methods more generally \citep{DBLP:journals/tacl/JacoviG21}.
In this paper, 
we present a systematic framework to characterise FIDAMs, and derive several new FIDAMs based on that framework. 
We then proceed with creating an evaluation pipeline that measures a FIDAM's ability to recover the structural rules for which we have good evidence that they play an important role in the target model's performance (Figure \ref{fig:pipeline}).
We first test this on a set of small-scale formal language tasks, that provide stronger faithfulness guarantees.
Finally, we present a case study of a large language model on the CoLA task for linguistic acceptability.

We find that the performance of FIDAMs is very variable, and that the performance on the small-scale formal language tasks may not be predictive of the performance of methods on the large-scale natural language task. 
This is an illustration of what we call the {\bf Attribution Generalisation problem}. 
We argue that this problem remains a key open problem in the study of explanation methods in general. 

\section{Related Work: Assessing Faithfulness}\label{sec:related}

In this section we discuss related work on assessing the faithfulness of feature attribution methods (FAMs). 
A model explanation ideally provides better insights into model behaviour. However, it is important that an explanation is faithful to the reasoning of the model, and not merely plausible to a researcher. Unfortunately, attribution models can yield vastly different outcomes \citep{DBLP:journals/corr/abs-2205-04559}. 

Defining a notion of faithfulness itself is an ongoing debate, and it has been argued that we should not be aiming for a binary notion, but a graded one instead \citep{DBLP:journals/tacl/JacoviG21}.
To this end, various methodologies have been proposed to evaluate the faithfulness of explanation methods.

One research direction introduces metrics to evaluate faithfulness by quantifying the impact of features that were deemed to contribute the most by an attribution method.
\citet{DBLP:conf/nips/HookerEKK19} does this by \textit{retraining} a model on data from which the most contributing features have been removed.
\citet{DBLP:conf/acl/DeYoungJRLXSW20} provide a more direct measure, by quantifying changes in model predictions when only a subset of the most contributing features is fed to model.
\citet{DBLP:conf/emnlp/AtanasovaSLA20} build on this notion, introducing a range of diagnostic metrics that capture various aspects of explanation quality including faithfulness, human rationale agreement, and explanation consistency.
\citet{jain-etal-2020-learning} ensure and evaluate faithfulness by only allowing a model access to the set of features that were deemed important by the explanation method, which has also been shown to improve model robustness \citep{wiegreffe-etal-2021-measuring, DBLP:journals/corr/abs-2210-13575}.

Another line of work modifies the training data in such a way that we obtain guarantees of certain features the model must be paying attention to when making a prediction: e.g. by shuffling test data such that only part of the input resembles the statistics from the train set  \citep{DBLP:conf/acl/SchutzeRP18}, or by explicitly adding exploitable heuristics in the train set \citep{DBLP:journals/corr/abs-2111-07367, DBLP:conf/iclr/AdebayoMAK22}.
These two approaches could be characterised as \textit{grey box} models: we adapt the data in such a way that we gain a degree of confidence what cues the model must be relying on, without having a full understanding of the model's internal reasoning.
A \textit{glass box} model, on the other hand, is a model whose behaviour is fully understood: it's not derived by training a model on a task, but hand-crafted.
\citet{hao-2020-evaluating} utilises such models to evaluate FAMs on formal language tasks, providing more robust guarantees on model behaviour. 

Our own approach is related to the first line of research, making use of \textit{grey box} models.
Instead of evaluating FAMS, we evaluate FIDAMs, that provide more comprehensive insights into model reasoning.
Deployment of such methods within NLP has been fairly limited, and as such evaluating their faithfulness in a language context has been an underexplored research topic.


 


\section{A Framework for Characterising FIDAMs}
%
Feature attribution methods typically decompose a model prediction into a sum of feature contributions \citep{DBLP:conf/icml/SundararajanTY17, DBLP:conf/nips/LundbergL17}.
A large contribution then indicates that this feature played an important role in a model's prediction.
Although feature attributions can provide meaningful insights into the inner model dynamics, they paint a fairly limited picture of the model behaviour.
Most importantly, \textbf{interactions} between features are lumped together, making it impossible to discern whether a large contribution of a feature stemmed from that feature alone, or from its interaction with neighbouring features.
To address this, multiple methods have been proposed that decompose a model prediction into a sum of feature interactions, based on similar mathematical formalism as those of feature attributions.

\paragraph{Notation}
A neural network is represented as a single function $f$.
    The input to $f$ is denoted as $\mathbf{x}$, which consists of $N$ input features.
A partial input $\mathbf{x}_S$ only consists of input features $S\subseteq N$.
A value function $v(\mathbf{x}_S)$ quantifies the model output on the partial input $\mathbf{x}_S$.
Padding the missing features in $\mathbf{x}_S$ with replacement features $\mathbf{x}_{\setminus S}'$ is denoted as $\mathbf{x}_S \cup\mathbf{x}_{\setminus S}'$.
The attribution value of feature $i$ is denoted as $\phi_i$, and the interaction effect of a set of features $\mathcal{I}$ is denoted as $\Gamma_\mathcal{I}$.

\paragraph{Attribution Dimensions}

Attribution methods can generally be characterised along two dimensions \citep{DBLP:journals/jmlr/CovertLL21}: 1) how the method deals with feature removal, and 2) how the impact of removing a feature is quantified.
FIDAMs are built on the same principles as FAMs, and can be categorised along the same two dimension.
By discerning these two dimensions we can separately evaluate their impact on the faithfulness of the attribution method.
Furthermore, we can combine feature removal procedures with influence quantification methods in order to obtain novel attribution methods, an observation that has also been made in the context of FIDAMs by \citet{jiang2023weighted}, who, concurrent to our work, provide a general framework for characterising FIDAMs. 


\subsection{Feature Removal}\label{sec:baselines}
It is not straight-forward to define the absence of a feature to a model's input.
The main goal here is to replace the removed feature with a neutral \textbf{baseline}, that adequately represents the absence of the feature.
Methods often make use of a neutral input feature, the \textbf{static baseline} $\mathbf{x}'$, such as a zero-valued embedding or a pad token:
\begin{equation}\label{eq:static_baseline}
    v(\mathbf{x}_S) = f(\mathbf{x}_S \cup \mathbf{x}'_{\setminus S})
\end{equation}
This may, however, lead to input that lies outside of the original input distribution \citep{kim-etal-2020-interpretation}.
The reason why this is problematic is that the model may behave erratically on such modified input, posing issues to the faithfulness of the explanation.

Instead of using a static baseline, we can also opt to use a baseline that is sampled from a \textit{background distribution} \citep{DBLP:conf/sp/DattaSZ16}.
There exist two approaches to this procedure \citep{DBLP:conf/icml/SundararajanN20, DBLP:journals/corr/abs-2006-16234}.
The \textbf{observational conditional expectation} samples the baseline features from a distribution that is conditioned on the set of features that are still present in the input \citep{DBLP:conf/nips/FryeRF20, DBLP:journals/ai/AasJL21}:
\begin{equation}\label{eq:conditional_expection}
    v(\mathbf{x}_S) = \mathbb{E}_{\mathbf{x}'_{\setminus S}}\left[f(\mathbf{x}_S \cup \mathbf{x}'_{\setminus S})~|~\mathbf{x}_S\right]
\end{equation}
The \textbf{interventional conditional expectation} drops the conditional, and samples the baseline features from an independent distribution:
\begin{equation}\label{eq:interventional_expectation}
    v(\mathbf{x}_S) = \mathbb{E}_{\mathbf{x}'_{\setminus S}}\left[f(\mathbf{x}_S \cup \mathbf{x}'_{\setminus S})\right]
\end{equation}
There exist two motivations for the latter approach: \citet{DBLP:conf/nips/LundbergL17} drop the conditional expectation for computational reasons, allowing them to approximate the observational conditional expectation. 
\citet{DBLP:conf/aistats/JanzingMB20} provide a perspective derived from causality theory, stating that the \textit{intervention} of removing a feature should break the dependence between the baseline and remaining features, and hence conditioning on these features is fundamentally wrong.

The previous two methods sample baseline values for individual missing features, but we can also compute the expectation over the range of possible baselines.
This yields the technique of \textbf{expected explanations} \citep{DBLP:journals/natmi/ErionJSLL21}, in which attributions with different static baselines are averaged out over a background distribution $D$:
\begin{equation}
    \phi_i = \mathbb{E}_{\mathbf{x}'\thicksim D}\left[\phi_i(\mathbf{x};\mathbf{x}') \right]
\end{equation}

\subsection{Quantifying Feature Influence}\label{sec:influence}
The simplest method of quantifying the influence of a feature is expressed as the output difference after \textbf{ablating} the feature:
\begin{equation}\label{eq:ablation}
    \phi_i = v(\mathbf{x}) - v(\mathbf{x}_{\setminus i})
\end{equation}
Note that this formulation can be combined with any of the feature removal methods: e.g. Occlusion \citep{DBLP:conf/eccv/ZeilerF14} combines this influence method with a static baseline (Eq. \ref{eq:static_baseline}), whereas \citet{kim-etal-2020-interpretation} combines it with the observational conditional expectation \mbox{(Eq. \ref{eq:conditional_expection})}, employing BERT as the conditional distribution.

A more involved method leverages a technique from the field of game theory, called the \textbf{Shapley value} \citep{shapley1953value}.
Shapley values were originally introduced in the domain of cooperative games, in which players can form coalitions to change the outcome of the game.
This setup can be transferred directly to machine learning models, in which features now take up the role of the players.
A Shapley value expresses the contribution of a feature as the marginal gain of including that feature in the input, averaged over all possible coalitions of features.

\section{FIDAMs}\label{sec:interactions}
We now address a series of interaction methods that we use in our own experiments.


\paragraph{Group Ablation}
The feature influence principle of Equation \ref{eq:ablation} can straightforwardly be extended to \textit{groups} of features. 
In our experiments we will focus on pairwise interactions, but any kind of feature subset can be used here.
\begin{equation}
    \Gamma_{i,j} = v(\mathbf{x}) - v(\mathbf{x}_{\setminus ij})
\end{equation}

\paragraph{Archipelago}
Explaining model behaviour in terms of pairwise interactions will already yield a better portrayal of its internal behaviour than `flat' attributions, but it neglects the interactions that occur within larger groups of features. 
Archipelago \citep{DBLP:conf/nips/TsangR020} splits up the feature interaction procedure into two phases: first an interaction detection method is performed that clusters features into interaction sets, and afterwards interaction scores are assigned to these sets as a whole.
Interaction detection is based on measuring the non-additive effect of pairs of features.
The interaction effect that is assigned to an interaction set $\mathcal{I}$ is expressed as follows, with respect to a static baseline $\mathbf{x}'$:
\begin{equation}
    \Gamma_\mathcal{I} = f(\mathbf{x}_\mathcal{I} \cup \mathbf{x}_{\setminus\mathcal{I}}') - f(\mathbf{x}')
\end{equation}
Note that Archipelago expresses the interaction effect inversely compared to the Group Ablation procedure: instead of measuring the impact of removing a group of features, we now measure the impact of solely keeping this group in the input.

\paragraph{Shapley(-Taylor) Interaction Index}
Both the previous methods base interaction effects on direct output differences.
We can modify the formulation of the Shapley value to yield interaction effects.
This modification was originally introduced in the field of game theory, called the Shapley Interaction Index \citep[SII,][]{10.2307/2661445, DBLP:journals/ijgt/GrabischR99}.
Instead of computing the marginal gain that is achieved by a single feature, we now compute the marginal gain of \textit{groups} of features.
The Shapley-Taylor Interaction Index \citep[STII,][]{DBLP:conf/icml/SundararajanDA20} is an extension of SII, satisfying additional theoretical properties.

\paragraph{Hessian}
Analogous to utilising the gradient for feature attributions, we can employ the second-order derivative to quantify interactions between features, which is captured by the Hessian matrix.
\citet{friedman2008predictive} and \citet{DBLP:conf/icml/SorokinaCRF08} consider an interaction between two variables to exist when the effect of one variable on the response depends on values of the other variable, which can be expressed in terms of the second-order partial derivative:
\[\Gamma_{i,j} = \left[\frac{\partial^2f(\mathbf{x})}{\partial x_i\partial x_j}\right]^2\]
A common approach when using the gradient of a model as a proxy for feature importance is to multiply it with the input embeddings \citep{DBLP:conf/icml/ShrikumarGK17, DBLP:series/lncs/AnconaCOG19}: in our experiments we consider an analogous method to the Hessian that we call \textbf{Hessian $\times$ Input}. 

\paragraph{Integrated Hessians}
Directly using the Hessian as explanation method is prone to the same caveats as using the gradient: the interactions signal may vanish due to saturation.
Integrated Hessians \citep[IH,][]{DBLP:journals/jmlr/JanizekSL21} address this issue by integrating over the Hessian manifold along a path between the input and a baseline.
This is achieved by applying the method of Integrated Gradients \citep{DBLP:conf/icml/SundararajanTY17} to itself.
An IH interaction between features $i$ and $j$ can hence be interpreted as the contribution of $i$ to the contribution of $j$ to the models prediction.
The path integral between input and baseline is approximated via a Riemann sum interpolation.

\paragraph{Other Methods}
The methods explained thus far have all been incorporated in our experimental pipeline.
The scope of our work focuses mainly on \textit{pairwise} interactions, but methods that extract higher-order interactions have been proposed as well \citep{DBLP:conf/iclr/JinWDXR20}. 
Comparing such methods to linguistic structure is an exciting avenue that we leave open to future work.
Other interaction methods that were not considered include two methods that preceded Archipelago: Neural Interaction Detection \citep{DBLP:conf/iclr/TsangC018} and MAHE \citep{DBLP:journals/corr/abs-1812-04801}.
The feature attribution method Contextual Decomposition \citep{DBLP:conf/iclr/MurdochLY18} has been extended to extract interactions as well \citep{DBLP:conf/iclr/SinghMY19, saphra-lopez-2020-lstms, DBLP:conf/acl/ChenZJ20}, but these methods place the constraint that only contiguous groups of features can interact.
Integrated Directional Gradients \citep{sikdar-etal-2021-integrated}, an extension of Integrated Gradients to capture \textit{group attributions}, could be adapted to our framework, but we leave this open for future work.

\section{Evaluating FIDAMs}
\label{sec:approach}
The final component of our framework is a methodology for evaluating the faithfulness of FIDAMs. 
To lay a robust foundation for such work, we propose to evaluate a range of interaction methods and baselines on smaller deep learning models (using LSTM and Transformer architectures) that have been trained to recognise formal languages, based on a probabilistic context-free grammar (PCFG). 

Our models are trained on a binary language classification task, in which a model needs to learn to discern between well-formed strings and minimally corrupted counterparts.
Models are trained to perfection (100\% accuracy) on both train and test set.
To obtain perfect performance, a model must rely solely on the grammatical rules that underlie the language, without resorting to spurious heuristics, because only these results allow completely solving the task.
This way, due to the controlled nature of the task, we obtain a high degree of confidence about the model's behaviour.

The goal of our experimental approach is to recover the structure of the language \textit{based on the trained model itself}.
This is achieved by the FIDAMs outlined in \S\ref{sec:interactions}.
We aim to uncover whether a structural dependency between two features results in a high interaction effect.
Since our models have been trained to perfection, this allows us to employ our setup as a way of measuring the \textbf{faithfulness} of a FIDAM.
A method that assigns a high interaction effect to features that contain a dependency in the original grammar is able to provide a faithful reflection of a model's understanding of the task.
By testing a wide range of FIDAMs and baselines we can uncover which configuration yields the most faithful explanations.
A graphical overview of our approach is depicted in Figure~\ref{fig:pipeline}.

\paragraph{Task}
The binary language classification task is set up by generating positive examples $D^+$, based on some PCFG, and negative examples $D^-$, derived from minimally corrupting the positive examples.
We split the union of these two sets into a random train/test split of 80/20\%.
We train our models with a default cross-entropy loss, using the AdamW optimiser \citep{DBLP:conf/iclr/LoshchilovH19}, a learning rate of 0.01, and a batch size of 48.

\paragraph{Models}
Our pipeline permits the use of any kind of neural model architecture, in our experiments we considered both LSTMs \citep{hochreiter97} and Transformers \citep{vaswani17}.
In our experiments we report the results of the LSTM model, but we observed similar results for Transformers: due to the black-box approach of our explanation procedure the architecture itself is not of great importance.
The models are deliberately small: we use an embedding size that is equal to the number of symbols in the language it is trained on, a hidden state size of 20, and a single layer.
This results in models that provide a compute-friendly test bed for evaluating the FIDAMs.

\begin{figure}
    \centering
    \includegraphics[width=0.75\columnwidth]{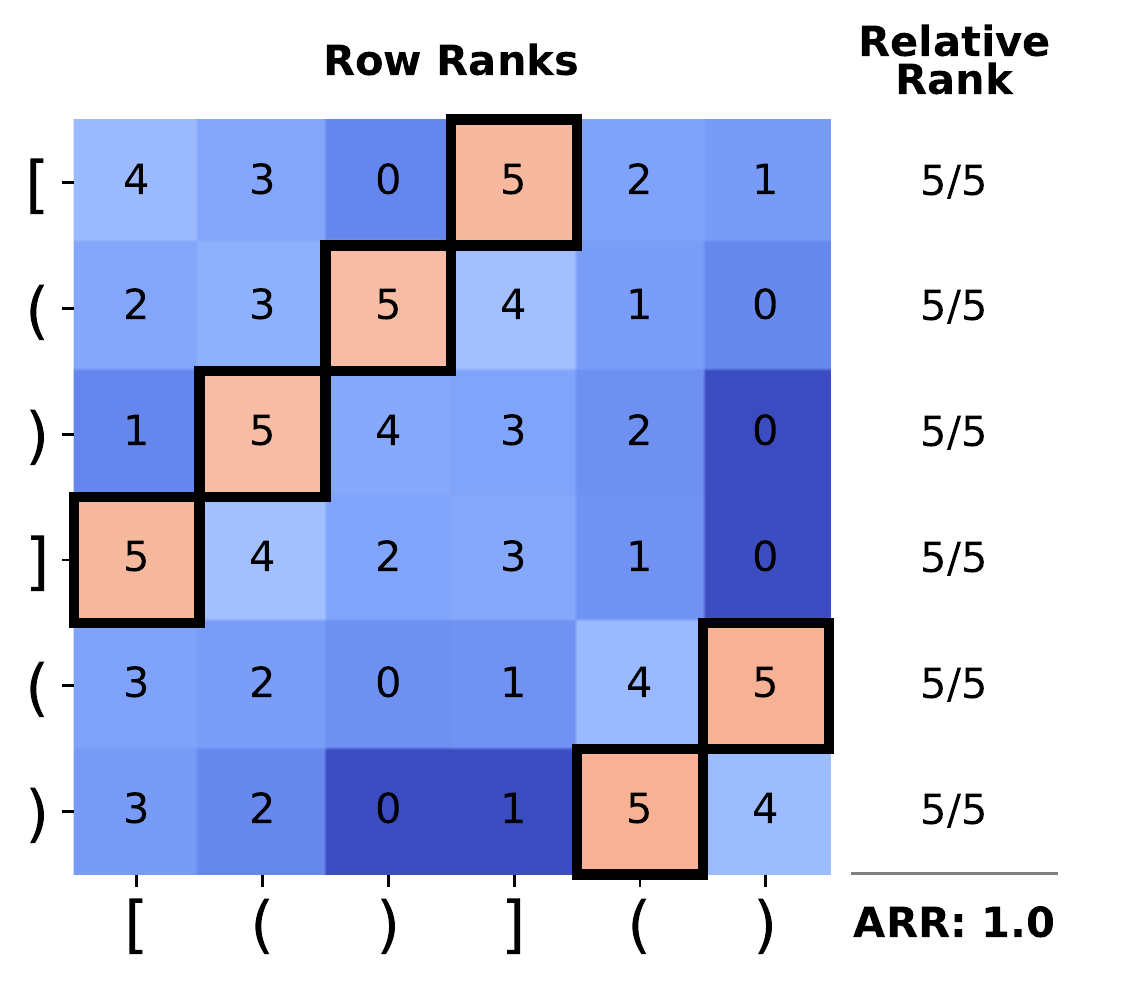}
    \caption{
    Example for the computation of the Average Relative Rank metric. For each row we compute the relative rank of the interaction of interest (here the Dyck language), and these row-wise relative ranks are averaged into a single score between 0 and 1.
    A random interaction matrix results in an ARR of around 0.5.
    }
    \label{fig:arr_example}
\end{figure}

\paragraph{Evaluation}
We focus on \textit{pairwise} interactions: interactions between individual pairs of features.
A FIDAM that extracts pairwise interactions for an input sequence $\mathbf{x}\in\mathbb{R}^{N}$ returns a matrix of interaction effects $\Gamma\in\mathbb{R}^{N\times N}$.
Since our goal is to uncover whether structural dependencies result in high interaction effects, we approach the evaluation of the interaction matrix as a retrieval task.
By aggregating and normalising the \textit{rank} of each interaction of interest we can quantify the performance of a FIDAM.
We call this metric the \textbf{Average Relative Rank} (ARR):
\begin{equation}\label{eq:arr}
    ARR(\Gamma, \mathcal{I}) = \frac{1}{|\mathcal{I}|} \sum_{i,j\in I} \frac{R(\Gamma_i)_j}{N-1}
\end{equation}
where $\mathcal{I}$ denotes the set of interaction pairs of interest and $R(\Gamma_i)$ denotes the rank of each interaction between feature $i$ and the other features in input $\mathbf{x}$ (the lowest interaction is ranked 0, and the highest interaction is ranked $N-1$).
We aggregate these scores over an evaluation set to obtain a general performance score of the FIDAM.
A graphical overview of this procedure is provided in Figure~\ref{fig:arr_example}.

\paragraph{Baselines}
We consider a range of baselines in our experiments, based on the procedures explained in \S\ref{sec:baselines}.
For the static baselines we consider a zero-valued baseline ($\mathbf{x}'=0$), and a baseline that utilises a fixed mapping $T$ based on the original input symbols ($\mathbf{x}'=T(\mathbf{x})$).
Expected attributions are marginalised over samples from the distribution of well-formed strings $D^+$ and corrupted strings $D^-$.
The interventional conditional expectation (Eq. \ref{eq:interventional_expectation}) is computed with a corpus-wide unigram distribution ($P(x_i)$), a unigram distribution that is conditioned on the sentence position ($P(x_i|i)$), and as a joint distribution over the missing features ($P(\mathbf{x}'_{\setminus S})$), that we sample from the training corpus.
The observational conditional expectation (Eq. \ref{eq:conditional_expection}) is computed based on the original corpus data.\footnote{Due to the small scale of the PCFGs considered here we can generated the complete language up to a certain length, and sample from strings that have feature overlap with the features that are still present in the partial input. For more complex tasks an auxiliary LM can be used instead.}  

\section{Experiments on Formal Languages}\label{sec:experiments}
We apply the evaluation procedure of \S\ref{sec:approach} to two formal languages: the Identity Rule language and the Dyck-2 language.
In the appendix (\S\ref{sec:palin}) we also present results on a palindrome language.

\subsection{Identity Rule}
The first language we consider is a regular language consisting of strings in which the first two symbols are identical, followed by a random sequence of symbols.
The language is formed by the following grammar:
\begin{lstlisting}
    S $\rightarrow$ $x$ $x$ A         $x\in\{a,b,c\}$
    A $\rightarrow$ $x$ A | $\epsilon$       $x\in\{a,b,c\}$
\end{lstlisting}
The only interaction of interest here is between the first two symbols; all subsequent symbols are irrelevant for the prediction.
An ARR score of 1.0 then indicates that for all corpus items the interaction between the first two items was the strongest out of all interactions.

We use a corpus size of 1.000, a maximum sequence length of 20, with 3 different input symbols.
Corrupted strings are derived by altering one of the first two symbols (e.g. \textit{\textbf{aa}bcb}~\textrightarrow~\textit{\textbf{\textcolor{red}{c}a}bcb}).

\paragraph{Results}
The results for an LSTM that was trained on the language are shown in Table~\ref{tab:identity_results}.
Due to the simplicity of the language and for brevity we only report results on three baselines.
A static zero-valued baseline provides imperfect interactions for all methods.
The Hessian, that does not depend on any baseline, performs better than all other methods here.
When sampling the baseline, however, multiple methods perfectly retrieve the interaction between the first two symbols for all corpus items.
Interestingly, Group Ablation and IH benefit from sampling from the distribution of well-formed items, whereas Archipelago performs best when sampling from the distribution of corrupted items.
    
\begin{table}[]
    \centering
    \small
    \begin{tabular}{l c c c c}
        & \textit{NB} & {\small$\mathbf{0}$} & {\small$\mathbf{x}'\thicksim D^+$} & {\small$\mathbf{x}'\thicksim D^-$}\\\midrule
 Group Ablation & -- & 0.49 &      \textbf{\textbf{1.00}} &      0.53 \\
    Archipelago & -- & 0.30 &      0.24 &      \textbf{\textbf{1.00}} \\
            SII & -- & 0.70 &      \textbf{\textbf{1.00}} &      \textbf{\textbf{1.00}} \\
           STII & -- & \textbf{0.83} &      \textbf{\textbf{1.00}} &      \textbf{\textbf{1.00}} \\
        Hessian &  \textbf{0.93} & -- &      -- &      -- \\
Hessian$\times$Input &  0.66 & -- &      -- &      -- \\
             IH & -- & 0.81 &      \textbf{\textbf{1.00}} &      0.31 \\
    \end{tabular}
    \caption{
    Average Relative Rank for the Identity Rule language, columns indicate different baseline procedures. An average rank of 1 indicates that the method (correctly) assigned the interaction between the first two tokens the highest score.
    \textit{NB} indicates these methods use no baseline.
    }
    \label{tab:identity_results}
\end{table}

\begin{table*}[t]
    \centering
    \small
    \setlength\tabcolsep{6pt}
    \begin{tabular}{l c c c@{\hspace{.5cm}}c c@{\hspace{.5cm}}c c c@{\hspace{.5cm}}c}
        & & \multicolumn{2}{L}{\textit{Static}} & \multicolumn{2}{L}{\textit{Expected}} & \multicolumn{3}{L}{\textit{Interventional}} & \multicolumn{1}{L}{\textit{Observational}} \\
        \cmidrule(r{\tbspace}){3-4} \cmidrule(r{\tbspace}){5-6} \cmidrule(r{\tbspace}){7-9} \cmidrule(r{\tbspace}){10-10}
         & \textit{No baseline} & {\small$\mathbf{0}$} 
         & {\small$T(\mathbf{x})$}
         & {\small$D^+$} 
         & {\small$D^-$} 
         & {\small$P(x_i')$} 
         & {\small$P(x_i'|i)$}
         & {\small$P(\mathbf{x}'_{\setminus S})$}
         & {\small$P(\mathbf{x}'_{\setminus S}|\mathbf{x}_S)$} 
         \\\midrule
 Group Ablation & -- & \textbf{0.684} &   \textbf{1.000} &     0.916 &     0.884 &                       0.822 &                      0.821 &                 0.938 &        0.956 \\
    Archipelago & -- &0.466 &   0.528 &     0.250 &     0.554 &                         -- &                        -- &                   -- &          -- \\
            SII & -- & 0.555 &   \textbf{1.000} &     \textbf{0.921} &     \textbf{0.895} &                       0.876 &                      0.885 &                 0.923 &        0.989 \\
           STII & -- &0.583 &   0.999 &     0.876 &     0.820 &                       \textbf{0.881} &                      \textbf{0.906} &                 \textbf{0.952} &        \textbf{0.991} \\
        Hessian & 0.413 & -- &     -- &       -- &       -- &                         -- &                        -- &                   -- &          -- \\
Hessian$\times$Input & \textbf{0.542} & -- &    -- &       -- &       -- &                         -- &                        -- &                   -- &          -- \\
             IH & -- & 0.591 &   0.837 &     0.723 &     0.665 &                         -- &                        -- &                   -- &          -- \\

    \end{tabular}

    \caption{Average Relative Ranks for the Dyck language (higher indicates stronger alignment with Dyck rules), columns indicate different baseline procedures. }
    \label{tab:dyck_results}
\end{table*}

\subsection{Dyck-2}
The Dyck language is the language of well-nested brackets, and is a popular testbed for research on formal languages.
It is a context-free language with center embedding clauses, requiring a model to keep track of a memory stack while processing a string.
Earlier work on Dyck languages has shown that a wide range of neural model architectures can learn the grammar, including LSTMs \citep{DBLP:conf/emnlp/SennhauserB18}, memory augmented RNNs \citep{DBLP:journals/corr/abs-1911-03329}, Transformers \citep{ebrahimi-etal-2020-self}, and handcrafted RNNs \citep{hewitt-etal-2020-rnns, hao-2020-evaluating}.
We consider the Dyck-2 language, consisting of two types of brackets.
The language is formed by the following grammar:
\begin{lstlisting}
    S $\rightarrow$ $[$ S $]$ | $($ S $)$ | S S | $\epsilon$
\end{lstlisting}

We use a corpus size of 15.000, a maximum sequence length of 20, and a maximum branching depth of 4.
We use the same branching probabilities as \citet{DBLP:journals/corr/abs-1911-03329}, which results in a uniform probability of 0.25 for each rule.
Corrupted strings are derived by flipping a single bracket to any other bracket.
For the baseline mapping $T(\mathbf{x})$, we map a bracket to the other bracket type, i.e. `(' $\leftrightarrow$ `[' and `)'  $\leftrightarrow$ `]'.
This results in a baseline that is of the same structure as the original input, but without feature overlap.

\paragraph{Results}
We report the results for this language in Table~\ref{tab:dyck_results}, computed over all our baselines for an LSTM.
The zero-valued baseline again turns out to be a mediocre baseline: for none of the methods this results in a high ARR score.
The method that performs best is the fixed mapping $T(\mathbf{x})$.
For Group Ablation, SII, and STII this results in a perfect ARR; for IH it is the best performing baseline.

It is encouraging that a baseline exists that results in perfect ARR scores, but this mapping depends strongly on the nature of the Dyck task itself.
It is, for example, unclear how this static mapping would transfer to the natural language domain.
Ideally, a more general solution makes no strong assumptions about the baseline itself.
The three other baseline types in Table~\ref{tab:dyck_results} may provide such a solution, as these only depend on the access to the original training data.
Out of these, the observational baseline performs best: for the SII and STII methods this baseline performs nearly on par with the static mapping.
Obtaining this conditional distribution is challenging for more complex tasks, and it can be seen here that the interventional baseline with a joint distribution over the missing features performs well too.


\section{A Natural Language Case Study: CoLA}\label{sec:llm_experiments}
As a case study on a larger scale natural language task, we apply our methodology to language models fine-tuned on the CoLA task \citep{DBLP:journals/tacl/WarstadtSB19}.
CoLA is part of the GLUE Benchmark \citep{DBLP:conf/iclr/WangSMHLB19}, and is defined as a binary classification task of determining the linguistic acceptability of a single input sentence.
The task consists of linguistically valid sentences, and sentences that contain either a syntactic, semantic, or morphological violation.
A model that performs well on this task must have a thorough grasp of grammatical structure, and as such it provides a useful test bed for our FIDAM evaluation procedure.

In the previous experiments there was a degree of certainty about the structure that must be encoded by the model.
In the natural language domain, however, we do not have such certainty, and should therefore be careful of making strong claims about faithfulness.
Furthermore, natural language is highly multi-faceted and can not be captured by a single hierarchical structure that covers all these facets.
Nonetheless, we consider it valuable to test our setup on a natural domain in order to see if interesting differences between FIDAMs arise, and whether particular facets of language such as syntactic dependency structure can be extracted.

\subsection{Experimental Setup}
For our experiment we consider the RoBERTa-base model \citep{DBLP:journals/corr/abs-1907-11692} which obtains a Matthew's Correlation Coefficient score of 69.70 on the in-domain validation split. 
We filter out sentences that contain words that are split into multiple subwords by the tokenizer, since this leads to issues with aligning the interactions of multiple subwords to the dependency graph that is used for evaluation.
Furthermore, we limit sentences to a max length of 14 in order to allow the STII and SII methods to be computed exactly without approximations.
This resulted in a subset of around 60\% of the original in-domain validation split that we will use in our experiment.

We evaluate the FIDAM scores on the dependency parse tree of the sentence, that we obtain with the parser of {\tt spaCy} \citep{Honnibal_spaCy_Industrial-strength_Natural_2020}.
The ARR score is computed based on the interaction of each token with its \textit{parent} token.
We omit the interaction of the token that has the \textsc{root} node as its parent.
An example of this procedure can be found in Appendix \ref{sec:example}.
Do note that our evaluation procedure is one of many possibilities: we make the assumption that a token should interact strongly with its parent, but other interactions are likely to play a role within the model as well. 
We leave a more detailed investigation into using different types of linguistic structure open for future work.

We again consider the FIDAMs of Group Ablation, STII/SII, and Integrated Hessians.
We leave out Archipelago, since its procedure of assigning features to a single interaction set is not feasible with our setup in which multiple child tokens might be interacting with the same parent token.
Due to computational constraints we were unable to compute the full Hessian matrix of the language model, whose computation scales quadratically in the number of input \textit{neurons} \citep[\S 5.4]{bishop2007}.
For the static baselines we again consider the zero-valued baseline, as well as the \texttt{<pad>} token.
The interventional baselines are obtained by computing simple count-based distributions over a sample of 100.000 sentences from the Google Books corpus.
The distributions are based on the tokenization of the model's tokenizer, and allow for computationally efficient sampling.
We leave the incorporation of an observational baseline for future work, where an auxiliary masked LM might provide a useful conditional probability distribution.

\subsection{Results}
The results for the experiment are shown in Table \ref{tab:cola_roberta}.
As expected, due to reasons outlined at the start of this section, none of the methods reaches ARR scores that are close to 1.
Nonetheless, it is encouraging to see that various method/baseline combinations attain ARR scores that are far above chance level, indicating that there exists a strong degree of alignment between feature interactions and dependency structure.
Contrary to the Dyck results, using a zero-valued baseline yields some of the highest ARR scores, which indicates that within RoBERTa's embedding space this baseline represents a better neutral value.

A closer inspection of these results shows that the ARR scores are strongly negatively correlated to sentence length: for Group Ablation with a \texttt{<pad>} baseline, for example, we obtain a Spearman correlation of -0.38 ($p<<0.001$, regression plot in Appendix \ref{sec:correlation}).
This is not surprising: as the sentence length increases, the chance of a token's largest interaction being with its parent decreases.
Another correlation of interest is between the ARR score and the model's prediction of a sentence's acceptability.
A high correlation would indicate that the FIDAM's alignment with dependency structure are indicative of a model's performance.
For this we obtain a Spearman correlation of 0.14 ($p=0.036$): a relatively weak result that indicates that the structure our FIDAM extracted is only partly driving the model's comprehension of the sentence structure.

\begin{table}[]
    \centering
    \small
    \begin{tabular}{l c c c c}
            & \multicolumn{2}{L}{\textit{Static}} & \multicolumn{2}{L}{\textit{Interventional}}  \\
        \cmidrule(r{\tbspace}){2-3} \cmidrule(r{\tbspace}){4-5} 
        & {\small$\mathbf{0}$} & {\small\tt <pad>} & $P(x_i')$ & $P(\mathbf{x}'_{\setminus S})$\\\midrule
 Group Ablation &  0.702 &  \textbf{0.757}  &  0.518  &  0.491 \\
           SII  &  \textbf{0.746} &  0.668  &  \textbf{0.714}  &  \textbf{0.696} \\
           STII &  0.741 &  0.708  &  0.704  &  0.658 \\
             IH &  0.577 &  0.516  &  --    & -- \\
    \end{tabular}
    \caption{
    Average Relative Ranks for the dependency tree recovery of RoBERTa fine-tuned on CoLA.
    }
    \label{tab:cola_roberta}
\end{table}

\section{Discussion \& Conclusions}\label{sec:discussion}
In this paper, we have presented a framework for characterising FIDAMs and evaluating their faithfulness.
For the characterisation we set out two dimensions, feature removal and feature influence, along which existing FIDAMs can be characterised, by extending the `Explaining by Removing' framework of \citeauthor{DBLP:journals/jmlr/CovertLL21} to also apply to FIDAMs. 
This allows us to place each of the known FIDAMs in a two-dimensional grid, and to define novel variants of these models.
As such, many of the methods that we incorporated in our experiments are novel FIDAMs, such as combining Archipelago with expected explanations and STII with an observational baseline.

To assess the faithfulness of FIDAMs, we made use of formal language theory and `grey box models'. 
We use formal grammars to generate multiple datasets, each with known feature interactions, and train deep learning models to perfection on those datasets. 
Using FIDAMs, we can then extract the learned feature interactions based on the model itself, and compare these interactions to the dependencies in the original grammar.
We demonstrate that only specific combinations of FIDAMs and baselines are able to retrieve the correct interactions, while methods such as Archipelago and Integrated Hessians consistently fail to do so.

Finally, we tested our methodology on a natural language case study using a model fine-tuned on the CoLA task for linguistic acceptability.
Our results on the formal language tasks either did not turn out to be predictive of this experiment or, alternatively, the results \textit{were} predictive but the LMs made less use of dependency graph information than we might have expected.
This illustrates the challenge of the Attribution Generalisation problem, and the open question remains how we can transfer faithfulness guarantees from a synthetic, controlled context to the domain of natural language and LLMs.

We do show, however, that under certain configurations feature interactions align to some degree with the (syntactic) dependency structure of a sentence.
This paves the way for revealing linguistic structure in a more direct way than, for instance, can be achieved with Structural Probes \citep{DBLP:conf/naacl/HewittM19}.
Investigating whether different methods and baseline configurations are able to retrieve different aspects of structure is an exciting next step that we look forward to exploring in more detail. 
This could be examined, for instance, through the lens of contrastive explanations \citet{DBLP:journals/corr/abs-2202-10419}, a procedure that demonstrates that different baselines can reveal different aspects of linguistic structure. 
Furthermore, investigating the role that attention plays in modelling interactions could be a fruitful line of work, for instance by incorporating \textit{context mixing} methods to our pipeline, such as \textit{Value Zeroing} \citep{mohebbi-etal-2023-quantifying} and \textit{ALTI} \citep{ferrando-etal-2022-measuring}.




\section{Limitations}
Our work has only considered \textit{pairwise} interactions, but linguistic structure can also manifest through higher-order interactions.
We show that our results on small-scale, formal languages, are different from our results on a natural language task.
It would be premature to conclude that small-scale, synthetic tasks can not be predictive of behaviour on more complex tasks, and a more detailed investigation into the properties of the task that play a role is a viable next step.
Some of the FIDAMs we considered, most notably SII and STII, are intractable for larger inputs (scaling $O(2^n)$), and a necessary step in employing these methods to larger models is to construct better approximation procedures, e.g. by adapting SHAP to SII as has been done before for tabular data by \citet{DBLP:journals/corr/abs-1802-03888}. 
More generally, although we believe our probabilistic formal language setup provides a important step forward, solving the Attribution Generalization problem -- i.e., showing that results for small setups generalize to very large model -- remains a key open problem.


\typeout{}
\bibliography{custom}
\bibliographystyle{acl_natbib}

\appendix
\clearpage
\section{Palindromes}\label{sec:palin}
One additional language we investigated is the context-free language of palindromes.
In order to process a palindrome, a model needs to keep track of the dependency between each token in the first half of the string with its counterpart in the second half.
Palindromes can contain a special symbol in the middle of a string to demarcate the two string halves, making it less ambiguous for the model at which point it should track whether the palindrome is well-formed.
In our experiments, however, we found our models to perform well on both forms of palindromes.
Furthermore, following \citet{DBLP:journals/corr/abs-1911-03329}, we use a homomorphic mapping $h$ for the second half of the string, allowing the model to use separate embeddings for symbols occurring in the first and second half of a string:
\begin{lstlisting}
  S $\rightarrow$ $x$ S $h(x)$ | $\epsilon$          $x\in\{a,b,c,\cdots\}$
\end{lstlisting}
We use a corpus size of 5.000, 10 different input symbols, and a maximum sequence length of 18.
For the fixed baseline mapping $T(\mathbf{x})$ we map a symbol onto another random symbol, preserving the grammaticality of the palindrome (e.g. \textit{abBA}~\textrightarrow~\textit{cdDC}).

\paragraph{Results}
The results for this language, trained with an LSTM, are shown in Figure~\ref{tab:palin_results}.
Again, the zero-valued baseline performs poorly, with most methods scoring ARRs even below chance level.
The fixed baseline mapping again performs well for Group Ablation, SII, and STII, although it is not the best performing baseline this time.
These three FIDAMs obtain perfect performance when using the expected baselines over a distribution of well-formed palindromes, which also holds for the interventional baseline with a joint distribution over the missing features.
This is in contrast to the Dyck results, where the observational baseline resulted in better ARR scores for all three of these methods.

\begin{table*}[]
    \centering
    \small
    \setlength\tabcolsep{6pt}
    \begin{tabular}{l c c@{\hspace{.5cm}}c c@{\hspace{.5cm}}c c c@{\hspace{.5cm}}c}
        & \multicolumn{2}{L}{\textit{Static}} & \multicolumn{2}{L}{\textit{Expected}} & \multicolumn{3}{L}{\textit{Interventional}} & \multicolumn{1}{L}{\textit{Observational}} \\
        \cmidrule(r{\tbspace}){2-3} \cmidrule(r{\tbspace}){4-5} \cmidrule(r{\tbspace}){6-8} \cmidrule(r{\tbspace}){9-9}
         & {\small$\mathbf{0}$} 
         & {\small$T(\mathbf{x})$}
         & {\small$D^+$} 
         & {\small$D^-$} 
         & {\small$P(x_i')$} 
         & {\small$P(x_i'|i)$}
         & {\small$P(\mathbf{x}'_{\setminus S})$}
         & {\small$P(\mathbf{x}'_{\setminus S}|\mathbf{x}_S)$} 
         \\\midrule
 Group Ablation & 0.450 &   \textbf{0.980} &     \textbf{1.000} &     \textbf{0.943} &                       {0.777} &                      \textbf{0.836} &                 \textbf{1.000} &        {0.939} \\
    Archipelago & 0.356 &   0.452 &     0.439 &     0.717 &                         -- &                        -- &                   -- &          -- \\
       SII & {0.472} &   {0.933} &     \textbf{1.000} &     0.892 &                       \textbf{0.804} &                      {0.817} &                 \textbf{1.000} &        \textbf{1.000} \\
  STII & {0.472} &   0.921 &     0.999 &     {0.917} &                       0.760 &                      0.792 &                 \textbf{1.000} &        0.999 \\
        Hessian & \textbf{0.523} &     -- &       -- &       -- &                         -- &                        -- &                   -- &          -- \\
Hessian$\times$Input & \textbf{0.523} &     -- &       -- &       -- &                         -- &                        -- &                   -- &          -- \\
             IH & 0.505 &   0.637 &     0.693 &     0.535 &                         -- &                        -- &                   -- &          -- \\

    \end{tabular}

    \caption{Average Relative Ranks for the palindrome language (higher is better). }
    \label{tab:palin_results}
\end{table*}

\section{ARR Example}\label{sec:example}
An example of a sentence with a high ARR (0.93), for the Group Ablation method with a \texttt{<pad>} baseline:

{\centering
\hspace{1.5cm}\includegraphics[width=.75\columnwidth]{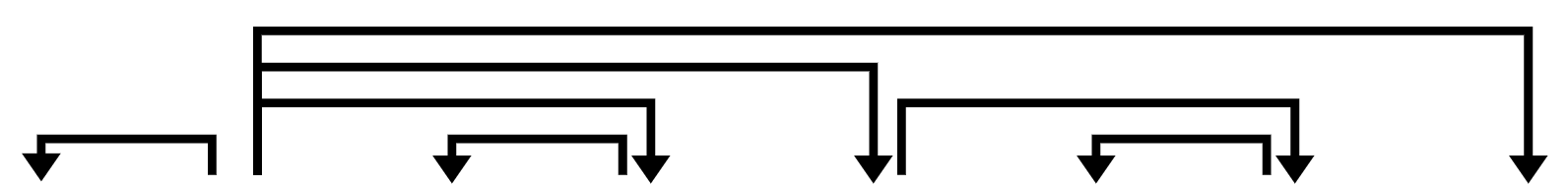}
\includegraphics[width=\columnwidth]{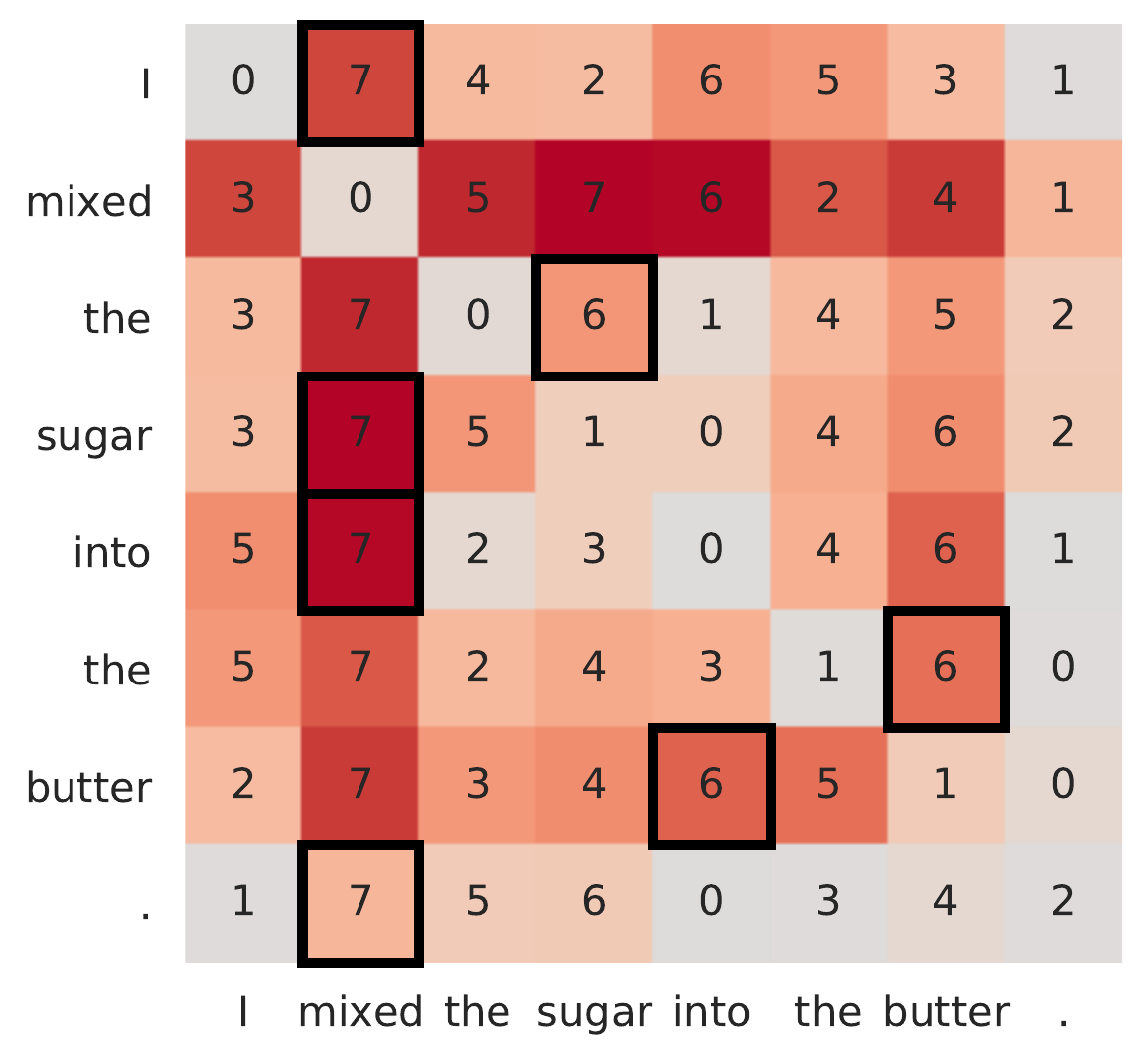}}

\section{Correlation CoLA ARR and sentence length}\label{sec:correlation}
Correlation between sentence length and ARR, shown here for Group Ablation with a \texttt{<pad>} baseline. Spearman's $\rho = -0.38$ ($p<<0.001$):

\includegraphics[width=\columnwidth]{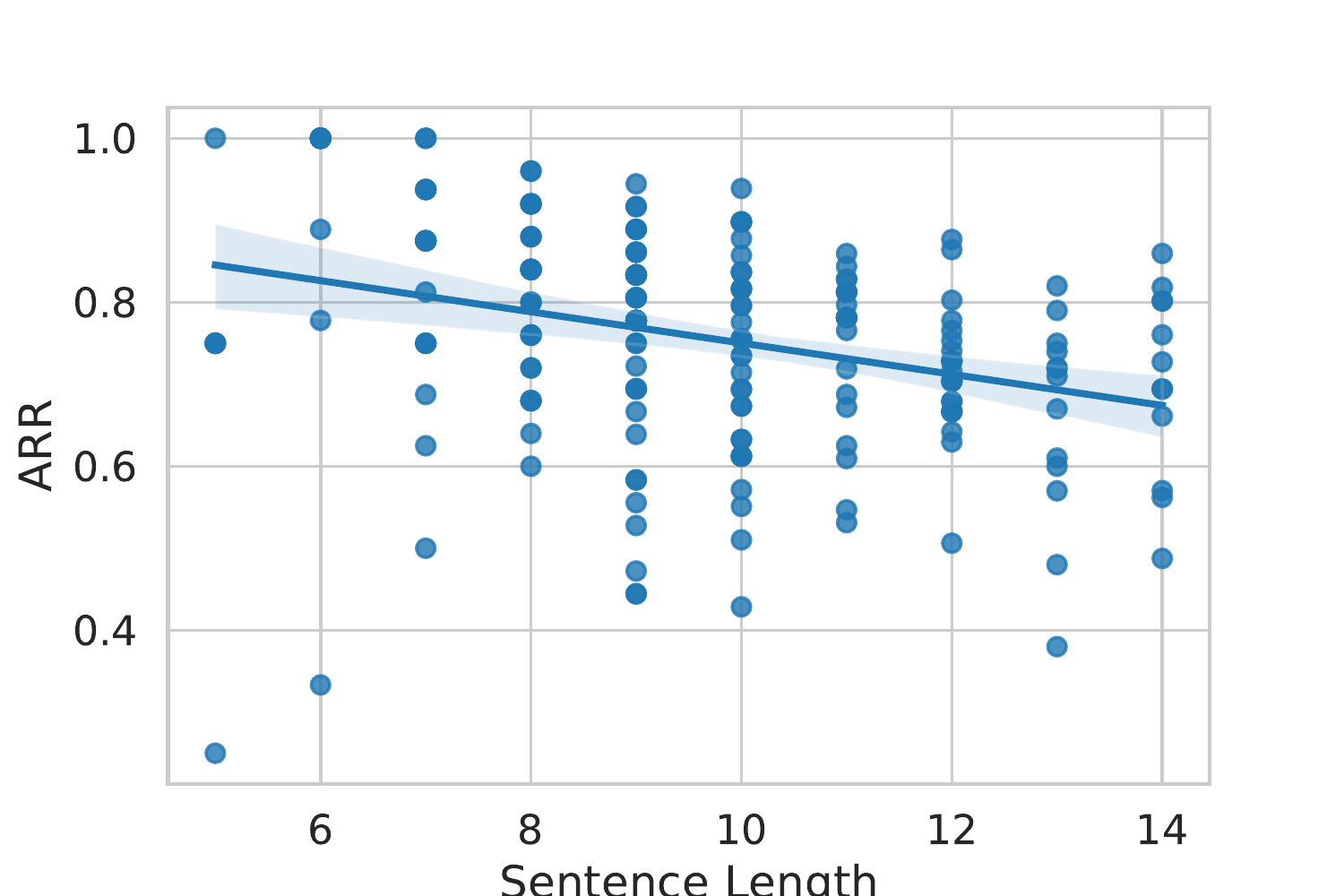}



\end{document}